\pdfoutput=1

\documentclass[11pt]{article}

\usepackage{latex/acl}

\usepackage{times}
\usepackage{latexsym}

\usepackage[normalem]{ulem}

\usepackage[T1]{fontenc}

\usepackage[utf8]{inputenc}

\usepackage{microtype}

\usepackage{inconsolata}

\usepackage{amssymb}
\usepackage{amsmath}
\usepackage{cleveref}
\usepackage{graphicx}
\usepackage{xcolor}

\usepackage{multirow}
\usepackage{tabularx}
\usepackage{booktabs}
\newcommand{\std}{\footnotesize}
\newcolumntype{Y}{>{\centering\arraybackslash}X}
\usepackage{enumitem}
\usepackage{changepage}

\usepackage{algorithm}
\usepackage{algpseudocode}

\title{Pre-trained Language Models Return Distinguishable Probability Distributions to Unfaithfully Hallucinated Texts}

\author{Taehun Cha \and Donghun Lee\thanks{\; corresponding author} \\
  Department of Mathematics \\
  Korea University \\
  \texttt{\{cth127, holy\}@korea.ac.kr} 
  }

\begin{document}
\maketitle
\begin{abstract}

In this work, we show the pre-trained language models return distinguishable generation probability and uncertainty distribution to unfaithfully hallucinated texts, regardless of their size and structure.
By examining 24 models on 6 data sets, we find out that 88-98\% of cases return statistically significantly distinguishable generation probability and uncertainty distributions.
Using this general phenomenon, we showcase a hallucination-reducing training algorithm.
Our algorithm outperforms other baselines by achieving higher faithfulness metrics while maintaining sound general text quality measures.\footnote{Source codes are available on \url{https://github.com/AIML-K/HalluDist}}

\end{abstract}

\section{Introduction}

\textit{Hallucination} is one of the key phenomena that undermine the reliability of large language models (LLMs), which recently gained large popularity in real-world applications \citep{zhang2023sirens}.
\citet{ji2023hallucination} characterized hallucinations with two perspectives: faithfulness and factuality.
The former represents consistency to the provided source text, while the latter is consistency to the world knowledge.
For example, if a user asks a machine to recommend a dinner menu and a machine answers that `Cereal is a breakfast menu enjoyed by many people', then the answer is factual but not faithful to the user's request.

Before the pre-trained language model (PLM) era, researchers found out that generation probability and uncertainty measured by a language model are correlated with the faithfulness of a text (\citealp{kang-hashimoto-2020-improved} and \citealp{xiao-wang-2021-hallucination}).
Though their work utilized un-pre-trained models trained on specific tasks, like image captioning, these works hinted at PLMs' potential to distinguish unfaithfulness.

In this paper, we examine three research hypotheses first.
(1) Does the unfaithfulness distinguishing ability generalize to the various sizes and types of PLMs?
(2) Does the model size affect the ability? 
(3) How does the fine-tuning affect the ability?
We examine these hypotheses with 24 pre-trained language models of various sizes and types on 6 data sets.
From massive experiments, 88-98\% cases return significantly distinguishable generation probability and uncertainty distributions.
Using this phenomenon, we showcase a simple training algorithm that effectively reduces hallucination.

\section{Related Works}
\label{sec:relworks}

\section*{Generation Probability/Uncertainty for Unfaithfulness Reduction}

While training, \citet{kang-hashimoto-2020-improved} reported truncating high-loss data points returns more faithful news titles.
The result implies training on a data point with a high loss (i.e. low generation probability) can make a model generate unfaithful texts.
While decoding, \citet{xiao-wang-2021-hallucination} showed that the model's predictive uncertainty shows a positive correlation with unfaithfulness in an image captioning task.
Though their work did not cover the PLMs, it hinted at the relationship between generation probability/uncertainty and faithfulness.
\citet{wan2023sequencelevel} extended this line of work when fine-tuning PLMs.
But their (un)certainty is computed from fine-tuned models, not a PLM itself, without verifying the fine-tuning effect.

\section*{LLM Probability/Uncertainty as a Factuality Measure}

As LLMs showed impressive performance on various tasks, hallucination researchers eagerly adopted LLMs in their works.
\citet{manakul2023selfcheckgpt} and \citet{azaria2023internal} reported LLMs' generation probability correlates well with factuality.
\citet{varshney2023stitch} utilized LLMs' generation probability to detect factually wrong texts.
However, their works concentrated on the factuality of generated texts, not faithfulness.
Moreover, they only utilized GPT3-like LLMs without verifying the size effect.
Our work focuses on faithfulness while verifying the size effect.

\section*{PLM as a Quality Measure}

PLMs can be used to construct quantitative metrics for various NLP tasks.
After \citet{Zhang*2020BERTScore:} and \citet{sellam-etal-2020-bleurt} introduced the BERTScore and BLEURT to compare generated and target texts, \citet{yuan2021bartscore} introduced the BARTScore to measure the generated text quality.
\citet{yuan2021bartscore} showed that BART's generation probability shows a positive correlation with various quality measures, like informativeness or coherence.
Our work is an extension and generalization of this work especially for the unfaithful hallucination.

\section{Suggested Metrics}
\label{sec:setup}

\subsection*{Notations}
Let $D=\{(x_i, y_i, h_i)\}_{i=1}^N$ be a data set.
For $i^{\text{th}}$ reference text $x_i$, let $y_i=(y_{i,1}, y_{i,2}, ... y_{i,n_i})$ be a corresponding target text, where $y_{i,j}$ represents $j^{\text{th}}$ token of $i^{\text{th}}$ target text.
Define $y_{i,!j}=(y_{i,1}, y_{i,2}, ..., y_{i,j-1}, [\text{MASK}], y_{i,j+1}, ..., y_{i,n_i})$, where a $j^{\text{th}}$ token is replaced with a [MASK] token, and $y_{i,<j}=(y_{i,1}, y_{i,2}, ..., y_{i,j-1})$, a truncated target text.
$h_i \in \{\text{Hallucinated, Entailed}\}$ is a unfaithful hallucination label.
If the content of $y_i$ is \textit{faithful} to the content of $x_i$, then $h_i=\text{Entailed}$.
On the other hand, if the content of $y_i$ is \textit{unfaithful} to the content of $x_i$, then $h_i=\text{Hallucinated}$.
For the convenience of notations, let $D_{\text{Hallucinated}}$ be a subset of the data set $D$ such that $h_i=\text{Hallucinated}$.
Likewise, define $D_{\text{Entailed}}$ similarly.

Let $f$ be a PLM.
$f$ can be an encoder model pre-trained on a masked language modeling task like BERT, a decoder model pre-trained on an autoregressive language modeling task like GPT2, or an encoder-decoder model like T5.
For an encoder model, $f(x_i, y_{i,!j})[v] \in [0,1]$ is a probability of a token $v$ at masked position $j$ given reference text and masked target text.
So $f(x_i, y_{i,!j})[y_{i,j}]$ is the probability for the right token.
For decoder and encoder-decoder models, $f(x_i, y_{i,<j})[v] \in [0,1]$ is a probability of a token $v$ at truncated position $j$ given reference text and truncated target text.
Hence $f(x_i, y_{i,<j})[y_{i,j}]$ is the probability for the right token.

\subsection*{Metrics}

Given a PLM $f$, we compute two metrics for each data point $(x_i, y_i)$.
These metrics, frequently used in the hallucination literature (\citealp{xiao-wang-2021-hallucination}, \citealp{manakul2023selfcheckgpt}, \citealp{varshney2023stitch} and \citealp{wan2023sequencelevel}), are as follow:

\begin{itemize}
    \item Log Token Probability (\textbf{LogProb}): A metric used to estimate the given model's generation probability of a target text. We compute the mean of log token probabilities of a target text by $\frac{1}{n_i} \sum_{j=1}^{n_i} \text{log} f(x_i, y_{i,<j})[y_{i,j}]$ when $f$ is either a decoder or an encoder-decoder model, and $\frac{1}{n_i} \sum_{j=1}^{n_i} \text{log} f(x_i, y_{i,!j})[y_{i,j}]$  when $f$ is an encoder model.
    \item \textbf{Entropy}: A metric frequently used to estimate the given model's prediction uncertainty of a target text. We compute the mean of entropy of each token for a target text by $\frac{1}{n_i} \sum_{j=1}^{n_i} \mathbb{E}_{v\sim f(x_i, y_{i,<j})}[-\text{log}f(x_i, y_{i,<j})[v]]$ when $f$ is either a decoder or an encoder-decoder model, and $\frac{1}{n_i} \sum_{j=1}^{n_i} \mathbb{E}_{v\sim f(x_i, y_{i,!j})}[-\text{log}f(x_i, y_{i,!j})[v]]$ when $f$ is an encoder model.
\end{itemize}

After computing the metrics for each data point, we obtain a metric distribution $\mathbb{P}$ for a data set.
With the $\mathbb{P}$, we can obtain an empirical cumulative distribution function (cdf), $F$.

\section{Distribution Distinguishability}
\label{sec:distinguishability}

Let $\mathbb{P}$ be a distribution of a metric based on a PLM $f$, and $\mathbb{D}$ be a statistic computing distinguishability between two distributions.
In this section, our goal is to verify (1) whether $\mathbb{D}(\mathbb{P}(D_{\text{Hallucinated}})||\mathbb{P}(D_{\text{Entailed}}))$ is statistically significant, and how the (2) model size and (3) fine-tuning of PLMs affect the distinguishability.

\begin{figure*}[t]
    \centering
    \includegraphics[width=1.0\linewidth]{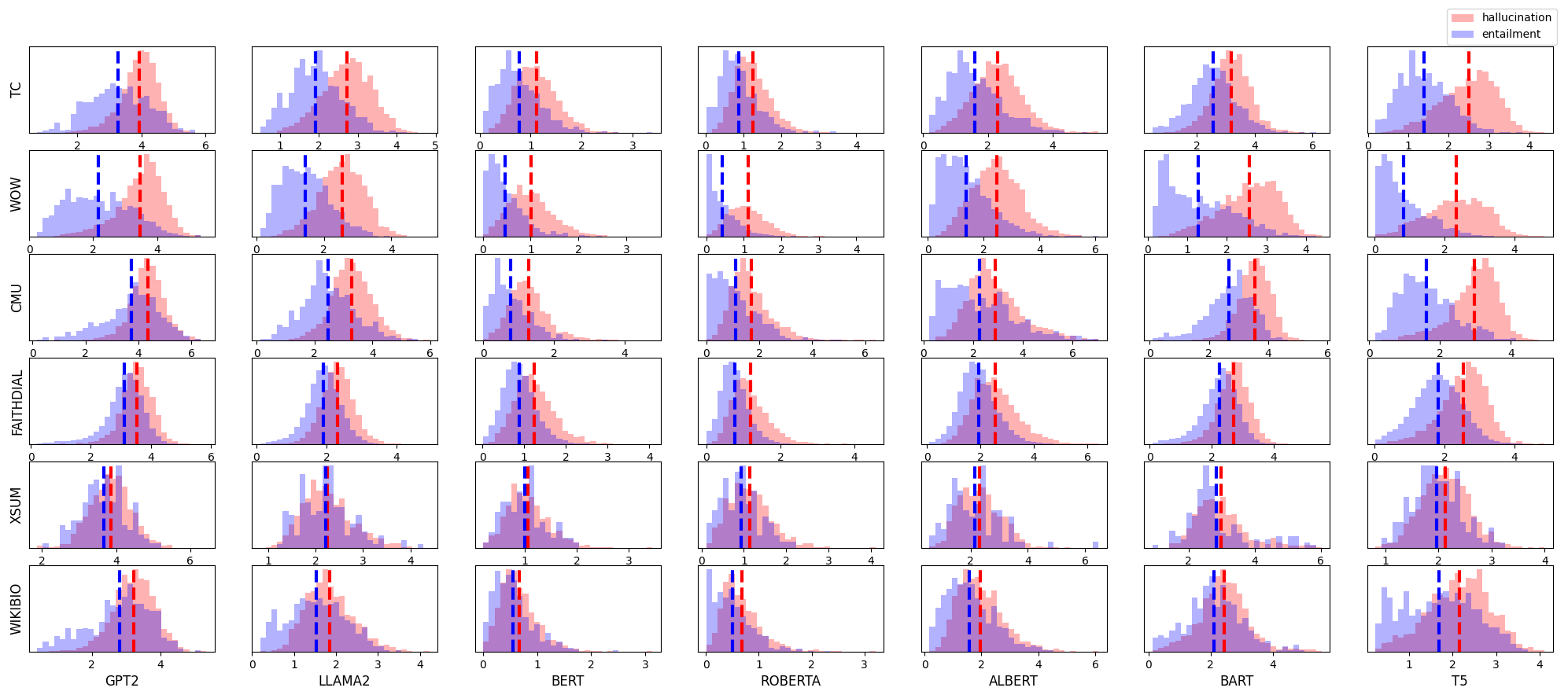}
    \caption{Empirical Entropy distribution and mean of $D_{Hallucinated}$ and $D_{Entailed}$ for each model and data set. We first compute Entropy for each data point, then separate the points according to the hallucination label. $x$-axis represents Entropy and $y$-axis represents the relative frequency. We plot the result of the smallest models for each model type.}
    \label{fig:type_entropy}
\end{figure*}

\subsection{Experimental Setup}

We utilize two statistics to quantify the distinguishability between two distributions.

\begin{itemize}
    \item Kolmogorov–Smirnov Statistic (\textbf{KS statistic}, \citealp{kolmogorov}): Given two cdfs, $F_1$ and $F_2$, the KS statistic is computed as $K(F_1, F_2)=\text{sup}_x|F_1(x) - F_2(x)|$. Intuitively, $K$ represents the maximum discrepancy between two cdfs. The KS statistic does not require distributional assumption unlike the t-test used in \citet{wan2023sequencelevel}.
    \item \textbf{Wasserstein Distance} \citep{kantorovich}: Given two one-dimensional cdfs, $F_1$ and $F_2$, the Wasserstein-1 distance is computed as $W(F_1, F_2)=\int_0^1|F_1^{-1}(q) - F_2^{-1}(q)| dq$, which represent the discrepant area between the two cdfs. 
\end{itemize}

\Cref{apdx:visualization} intuitively visualizes two metrics, given the two cdfs.

The KS statistic enables a non-parametric statistical test called the Kolmogorov–Smirnov test (KS test) on distributional differences, unlike the Wasserstein distance.
However, the KS statistic is sensitive to differences between the modes of two distributions while insensitive to their tails \citep{lipp2023wasserstein}.
We mainly utilize the KS statistic to test the significance of distinguishability and use the Wasserstein distance to compare the overall difference between models.


For a comprehensive analysis, we gather several natural language generation data sets containing hallucination label $h_i$ from multiple tasks.
For the knowledge-grounded dialogue task, we utilize BEGIN data set \citep{dziri-etal-2022-evaluating} and FaithDial data set (\textbf{FaithDial}, \citealp{dziri-etal-2022-faithdial}).
The BEGIN data set consists of three subsets based on existing data sets: TopicalChat (\textbf{TC}, \citealp{gopalakrishnan2019topical}), Wizard of Wikipedia (\textbf{WOW}, \citealp{dinan2019wizard}) and CMU Document Grounded Conversations (\textbf{CMU}, \citealp{cmu_dog_emnlp18}).
For the summarization task, we use XSum Hallucination Data Set (\textbf{XSum}, \citealp{maynez-etal-2020-faithfulness}), where human annotates unfaithful hallucination labels on machine-generated summaries.
For Wiki-like text generation, we utilize SelfCheckGPT data set (\textbf{WikiBio}, \citealp{manakul2023selfcheckgpt}) based on WikiBio data set \citep{lebret-etal-2016-neural}, where human annotates unfaithful hallucination labels on GPT-3 generated biography for corresponding Wikipedia page.
In summary, we utilize 6 data sets.
Basic statistics of each data are reported on \Cref{apdx:stat_hallu}.

Our analysis requires white-box models returning full probability distribution of tokens.
As a result, we utilize three general types of pre-trained open-source transformer models.
For the decoder model, we test 4 sizes of \textbf{GPT2} \citep{radford2019language} and 3 sizes of \textbf{Llama2} \citep{touvron2023llama}.
For the encoder model, we test 4 \textbf{BERT} \citep{devlin-etal-2019-bert}, 4 \textbf{ALBERT} \citep{Lan2020ALBERT:} and 2 \textbf{RoBERTa} \citep{liu2020roberta}.
For the encode-decoder model, we test 5 \textbf{T5} \citep{2020t5} and 2 \textbf{BART} \citep{lewis-etal-2020-bart}.
In summary, we test 24 models.


\subsection{Does PLM return Distinguishable Distributions to Unfaithful Texts?}
\label{subsec:distinguishable}

For each combination of a data set and a model, we implement the KS test on the KS statistics with a p-value of 0.01.
We compute the mean and standard deviation of the KS statistic.
The results are on \Cref{tab:distinguishable}.

\begin{table}[h]
    \centering
    \scalebox{0.85}{
    \begin{tabular}{c|c|c|c|c}
        \toprule
        & \multicolumn{2}{c|}{LogProb} & \multicolumn{2}{c}{Entropy} \\
        \cmidrule{2-5}
        & Sig. & KS & Sig. & KS \\
        \midrule
        \multirow{2}{*}{Encoder} & 88.33\% & 0.3144 & 90.00\% & 0.3274 \\
        & \std{(53 / 60)} & \std{(0.1097)} & \std{(54 / 60)} & \std{(0.1279)} \\
        \midrule
        \multirow{2}{*}{Decoder} & \textbf{92.86\%} & \textbf{0.3686} & 88.10\% & 0.3492 \\
        & \std{(39 / 42)} & \std{(0.1536)} & \std{(37 / 42)} & \std{(0.1589)} \\
        \midrule
        \multirow{2}{*}{Enc-Dec} & 88.10\% & 0.2652 & \textbf{97.62\%} & \textbf{0.3927} \\
        & \std{(37 / 42)} & \std{(0.1187)} & \std{(41 / 42)} & \std{(0.1698)} \\
        \bottomrule
    \end{tabular}
    }
    \caption{Summary table for the KS test. Sig. is a ratio of the statistically significant KS test with a p-value of 0.01. KS is the mean and standard deviation of the KS statistics.}
    \label{tab:distinguishable}
\end{table}

\textbf{Regardless of model type and metrics, PLMs return significantly distinguishable distributions for $D_{Hallucinated}$ and $D_{Entailed}$ for 88-98\% cases}.
For LogProb, decoder models return more significant results.
On the other hand, for Entropy, encoder-decoder models return more significant results.

Empirical Entropy distributions for each model and data set are presented on \Cref{fig:type_entropy}.
After computing an Entropy for each data point, we plot two histograms (in blue and red) with respect to the hallucination label.
Most cases return distinguishable mean and distributions.
PLMs tend to assign higher Entropy on $D_{Hallucination}$, representing higher uncertainty.
Meanwhile, PLMs tend to assign higher LogProb on $D_{Entailment}$ as shown in \Cref{apdx:logtokenprobability}.
Roughly speaking, PLMs are internally less confident and less certain when they predict hallucinated texts.

\subsection{Model Size Effect}
\label{subsec:size_effect}

\begin{figure}[h]
    \centering
    \includegraphics[width=1.0\linewidth]{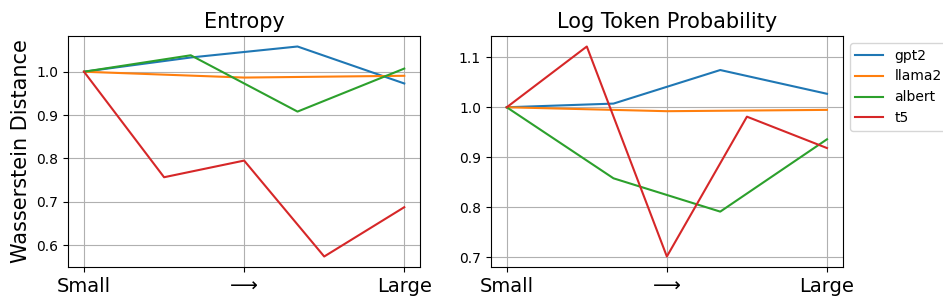}
    \caption{Visualization of the size effect. We divide all the Wasserstein distances with the distances from the smallest model to visualize the relative change as the size grows.}
    \label{fig:size_effect}
\end{figure}

Multiple researchers reported that GPT3-like LLMs can distinguish hallucinated texts (\citealp{manakul2023selfcheckgpt} and \citealp{azaria2023internal}).
It is natural to ask whether the distinguishing ability is enhanced as its size grows.
For comparison, we visualize Wasserstein distance of each metric relative to the smallest model's statistics.
To see the trend, we inspect models with more than three size variations, GPT2, Llama2, ALBERT, and T5.
We compute the mean of Wasserstein distance for all data sets.
The results are on \Cref{fig:size_effect}.

The results show that \textbf{bigger size does not guarantee better distinguishability}.
Notably for T5, the bigger model returns much less distinguishable distributions between hallucination and entailment groups.
This tendency is observed through all metrics.
Researchers should not blindly adopt LLMs to distinguish hallucinated texts without verifying their size effect.

\subsection{Fine-tuning Effect}
\label{subsec:training_effect}

Researchers utilized loss \citep{kang-hashimoto-2020-improved} or uncertainty \citep{xiao-wang-2021-hallucination} of the generation model once trained on target data to reduce unfaithfulness.
It is also natural to ask how the fine-tuning of PLM on target data affects the distinguishability.

\begin{figure}[h]
    \centering
    \includegraphics[width=1.0\linewidth]{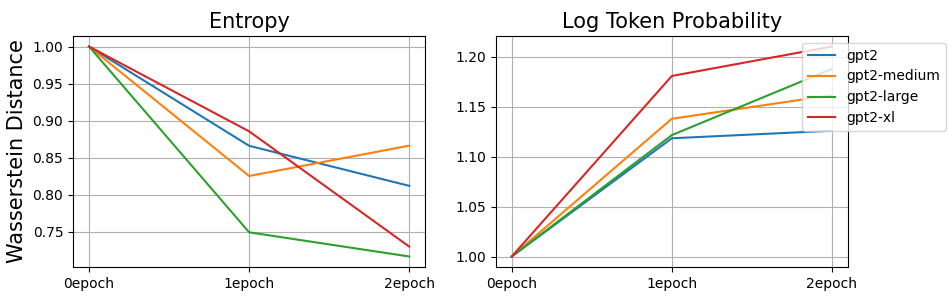}
    \caption{Fine-tuning effect for WOW data set. We divide all the Wasserstein distances with the distances from the pre-trained model to visualize the relative change as training proceeds.}
    \label{fig:training_effect}
\end{figure}

We train GPT2 models on the WOW and CMU training data set and check the statistics on the WOW and CMU portion of BEGIN data.
Note that the training data does not contain the portion from the BEGIN data set so data contamination does not occur.
The results for WOW data set are on \Cref{fig:training_effect}.

\textbf{The distinguishability from either metric is affected by fine-tuning while showing different trends}.
The distinguishability of LogProb increases as fine-tuning proceeds while the distinguishability of Entropy tends to decrease.
We find similar trends in the CMU data set, as shown in \Cref{apdx:cmu}.
Researchers should verify the fine-tuning effect of their target metric when they apply hallucination-reduction techniques.

\begin{table*}[t]
    \centering
    \scalebox{0.75}{
    \begin{tabular}{c|c|cccc|ccc}
    \toprule
        \multirow{2}{*}{Data Set} & \multirow{2}{*}{Method} & \multicolumn{2}{c}{$Q^2$} & \multirow{2}{*}{SummaC} & \multirow{2}{*}{FactKB} & \multirow{2}{*}{ROUGE-L} & BERT & BART \\
        ~ & ~ & F1 & NLI & ~ & ~ & ~ & Score & Score \\
        \midrule
        \multirow{6}{*}{WOW} & Unweighted & 0.6521 \std{(0.02)} & 0.6947 \std{(0.02)} & 0.2941 \std{(0.04)} & 0.5633 \std{(0.03)} & 0.2862 \std{(0.00)} & 0.3012 \std{(0.00)} & -2.7871 \std{(0.01)} \\
        ~ & CTRL & 0.6746 \std{(0.02)} & 0.7165 \std{(0.01)} & 0.3051 \std{(0.03)} & 0.5774 \std{(0.01)} & 0.2741 \std{(0.01)} & 0.3070 \std{(0.01)} & \underline{-2.7759} \std{(0.02)} \\
        ~ & Truncation & 0.6996 \std{(0.01)} & 0.7455 \std{(0.01)} & 0.4089 \std{(0.03)} & 0.6252 \std{(0.02)} & 0.2788 \std{(0.00)} & \underline{0.3133} \std{(0.00)} & -2.7998 \std{(0.02)}  \\ 
        ~ & mFACT & 0.7539 \std{(0.01)} & 0.7930 \std{(0.01)} & \underline{0.4988} \std{(0.04)} & 0.6966 \std{(0.03)} & \underline{0.3068} \std{(0.00)} & \textbf{0.3367} \std{(0.00)} & -2.8348 \std{(0.04)}  \\ 
        ~ & Ours-LogProb & \underline{0.7689} \std{(0.02)} & \underline{0.7946} \std{(0.02)} & 0.4287 \std{(0.04)} & \underline{0.7033} \std{(0.03)} & 0.2960 \std{(0.01)} & 0.2963 \std{(0.01)} & \textbf{-2.7633} \std{(0.05)}  \\ 
        ~ & Ours-Entropy & \textbf{0.7742} \std{(0.02)} & \textbf{0.8040} \std{(0.01)} & \textbf{0.5503} \std{(0.02)} & \textbf{0.7273} \std{(0.01)} & \textbf{0.3105} \std{(0.00)} & 0.3124 \std{(0.00)} & -2.7811 \std{(0.02)}  \\ 
        \midrule
        \multirow{6}{*}{FaithDial} & Unweighted & 0.7830 \std{(0.03)} & 0.8439 \std{(0.02)} & 0.1761 \std{(0.05)} & 0.6156 \std{(0.04)} & 0.3066 \std{(0.00)} & 0.3360 \std{(0.00)} & -2.7874 \std{(0.02)}  \\ 
        ~ & CTRL & 0.7758 \std{(0.01)} & 0.8405 \std{(0.01)} & 0.2255 \std{(0.05)} & 0.6267 \std{(0.04)} & 0.2921 \std{(0.00)} & 0.3384 \std{(0.00)} & -2.7769 \std{(0.04)}  \\ 
        ~ & Truncation & 0.7804 \std{(0.01)} & 0.8479 \std{(0.01)} & 0.3055 \std{(0.06)} & 0.6205 \std{(0.02)} & 0.2938 \std{(0.00)} & 0.3369 \std{(0.00)} & -2.7903 \std{(0.04)} \\ 
        ~ & mFACT & 0.8108 \std{(0.0)} & 0.8733 \std{(0.0)} & \textbf{0.4099} \std{(0.04)} & 0.6885 \std{(0.02)} & 0.3023 \std{(0.00)} & \textbf{0.3460} \std{(0.00)} & -2.8402 \std{(0.04)} \\ 
        ~ & Ours-LogProb & \textbf{0.8454} \std{(0.02)} & \underline{0.8841} \std{(0.02)} & 0.3652 \std{(0.10)} & \textbf{0.7706} \std{(0.04)} & \underline{0.3135} \std{(0.01)} & 0.3371 \std{(0.00)} & \underline{-2.7251} \std{(0.03)} \\
        ~ & Ours-Entropy & \underline{0.8403} \std{(0.02)} & \textbf{0.8905} \std{(0.01)} & \underline{0.4092} \std{(0.07)} & \underline{0.7475} \std{(0.03)} & \textbf{0.3179} \std{(0.00)} & \underline{0.3401} \std{(0.00)} & \textbf{-2.7166} \std{(0.02)} \\ 
        \midrule
        \multirow{6}{*}{MediQA} & Unweighted & 0.7912 \std{(0.01)} & 0.8333 \std{(0.01)} & 0.5152 \std{(0.02)} & \underline{0.9987} \std{(0.00)} & \underline{0.2491} \std{(0.01)} & 0.1712 \std{(0.01)} & -2.8650 \std{(0.03)} \\ 
        ~ & CTRL & 0.7754 \std{(0.02)} & 0.8189 \std{(0.02)} & 0.4899 \std{(0.02)} & \textbf{0.9988} \std{(0.00)} & 0.2355 \std{(0.01)} & 0.1602 \std{(0.01)} & -2.9055 \std{(0.04)} \\ 
        ~ & Truncation & 0.7784 \std{(0.01)} & 0.8180 \std{(0.01)} & \underline{0.5349} \std{(0.02)} & \textbf{0.9988} \std{(0.00)} & 0.2364 \std{(0.01)} & 0.1710 \std{(0.01)} & \textbf{-2.8126} \std{(0.05)} \\
        ~ & mFACT & \underline{0.7936} \std{(0.02)} & 0.8334 \std{(0.02)} & 0.5087 \std{(0.02)} & \textbf{0.9988} \std{(0.00)} & \textbf{0.2540} \std{(0.01)} & \textbf{0.1784} \std{(0.00)} & -2.8837 \std{(0.04)}  \\ 
        ~ & Ours-LogProb & \textbf{0.8129} \std{(0.02)} & \textbf{0.8579} \std{(0.02)} & \textbf{0.5416} \std{(0.01)} & 0.9927 \std{(0.01)} & 0.2447 \std{(0.01)} & \underline{0.1748} \std{(0.01)} & -2.8680 \std{(0.06)}  \\ 
        ~ & Ours-Entropy & 0.7853 \std{(0.02)} & \underline{0.8371} \std{(0.02)} & 0.4966 \std{(0.01)} & 0.9984 \std{(0.00)} & 0.2465 \std{(0.00)} & 0.1701 \std{(0.01)} & \underline{-2.8530} \std{(0.06)} \\ 
        \bottomrule
    \end{tabular}
    }
    \caption{Comparison table of faithfulness metrics (left) and text quality metrics (right). We mark the best score in bold and the second best with an underline.}
    \label{tab:weighted}
\end{table*}

\section{Hallucination Reduction with Weighted Training}
\label{sec:weighting}

In this section, we showcase a weighted training method to mitigate hallucination.
The idea is simple.
As we observe in \Cref{sec:distinguishability}, a data point with high Entropy tends to contain a hallucination.
Similarly, a data point with low LogProb tends to contain a hallucination.
Then what would happen if we use Entropy or LogProb as a loss weight for training?

We compare four baseline training methods on three data sets: the usual \textbf{Unweighted} training, a control-token method (\textbf{CTRL}, \citet{rashkin-etal-2021-increasing}), and other loss weighting methods (\textbf{Truncation} \citep{kang-hashimoto-2020-improved} truncate high loss points and \textbf{mFACT} \citep{qiu-etal-2023-detecting} weight the loss with faithfulness score).
We compare four faithfulness metrics and three general text quality metrics.
A more detailed explanation is in \Cref{apdx:experiment_detail}.
The results are on \Cref{tab:weighted}.

For knowledge-grounded dialogue data sets, our algorithms, with both Entropy and LogProb, improve faithfulness compared to Unweighted by a large margin, in all cases.
The same phenomenon occurs when compared with other baselines.
It is interesting since CTRL is designed to mitigate hallucination, especially in the knowledge-grounded dialogue task.

For MediQA, Ours-LogProb outperforms Unweighted on most metrics.
Ours-LogProb outperforms Truncation and mFACT on $Q^2$ and SummaC, though both are designed to reduce hallucination in the summarization task.
The result shows not only powerful hallucination reduction performance compared to other task-specific methods but also the general applicability of our methods through various tasks.

Our algorithm maintains general text quality measures, which we do not directly target.
More interestingly, our method often outperforms other baselines.
It may indicate its potential for enhancing not only faithfulness but also the overall fidelity and quality of generated text across diverse evaluation metrics.

\section{Conclusion}

Our work is the first comprehensive analysis of the PLMs' unfaithful hallucination-distinguishing ability.
We compare PLMs' generation probability and uncertainty distributions of unfaithful and entailed texts.
Regardless of the model type and size, PLMs return statistically distinguishable distributions to unfaithfully hallucinated and entailed texts for 88-98\% cases.
Unlike usual practice, the smaller models show comparable (and sometimes better) distinguishability to the largest models, while the distinguishability of Entropy declines while LogProb increases after fine-tuning.
Utilizing this phenomenon, we showcase a hallucination-reducing training algorithm that outperforms other baselines with hallucination reduction while maintaining sound general text quality measures.
We hope these findings lead to a deeper understanding of the hallucination phenomenon and more reliable hallucination mitigating techniques.

\section*{Limitation}

Though we made comparisons \textit{within} each model, comparison \textit{between} models raises a subtle issue.
For example, GPT2 and Llama2 are all decoder models but utilize different tokenizers.
The vocab size of GPT2 is 50,257 while Llama2 is 32,000.
As a result, the overall token probability is lower and entropy is higher for GPT2 since it should consider many more tokens for generation.
It makes cross-model comparison difficult and requires some generalized form of (un)certainty, which is beyond the scope of this work.

A PLM's unfaithful hallucination-distinguishing ability does not imply the faithfulness of the text generated from the PLM.
Likewise, confidence or certainty computed from the inner state of a model does not imply the certainty presented in the generated text.
As a result, a model can generate unfaithful text in a confident tone but with high entropy.
Researchers should not be confused between the tone and computed uncertainty especially when they work with black-box LLMs like GPT3.

As shown on \Cref{fig:type_entropy}, distinguishability is relatively vague for the XSum data set.
One possible reason can be the length of the reference text.
The mean length of reference text is 384 words, which is much longer than other data sets (at most 286 words).
Though we do not inspect the reference text length effect on entropy and log token probability, further analysis is required.

\section*{Acknowledgements}

This work is supported by the National Research Foundation of Korea (NRF) grant funded by the Korea government (MSIT) (No. 2020R1G1A1102828).

\bibliography{custom}

\appendix

\section{Visualization of KS Statistic and Wasserstein Distance}
\label{apdx:visualization}

\begin{figure}[h]
    \centering
    \includegraphics[width=1.0\linewidth]{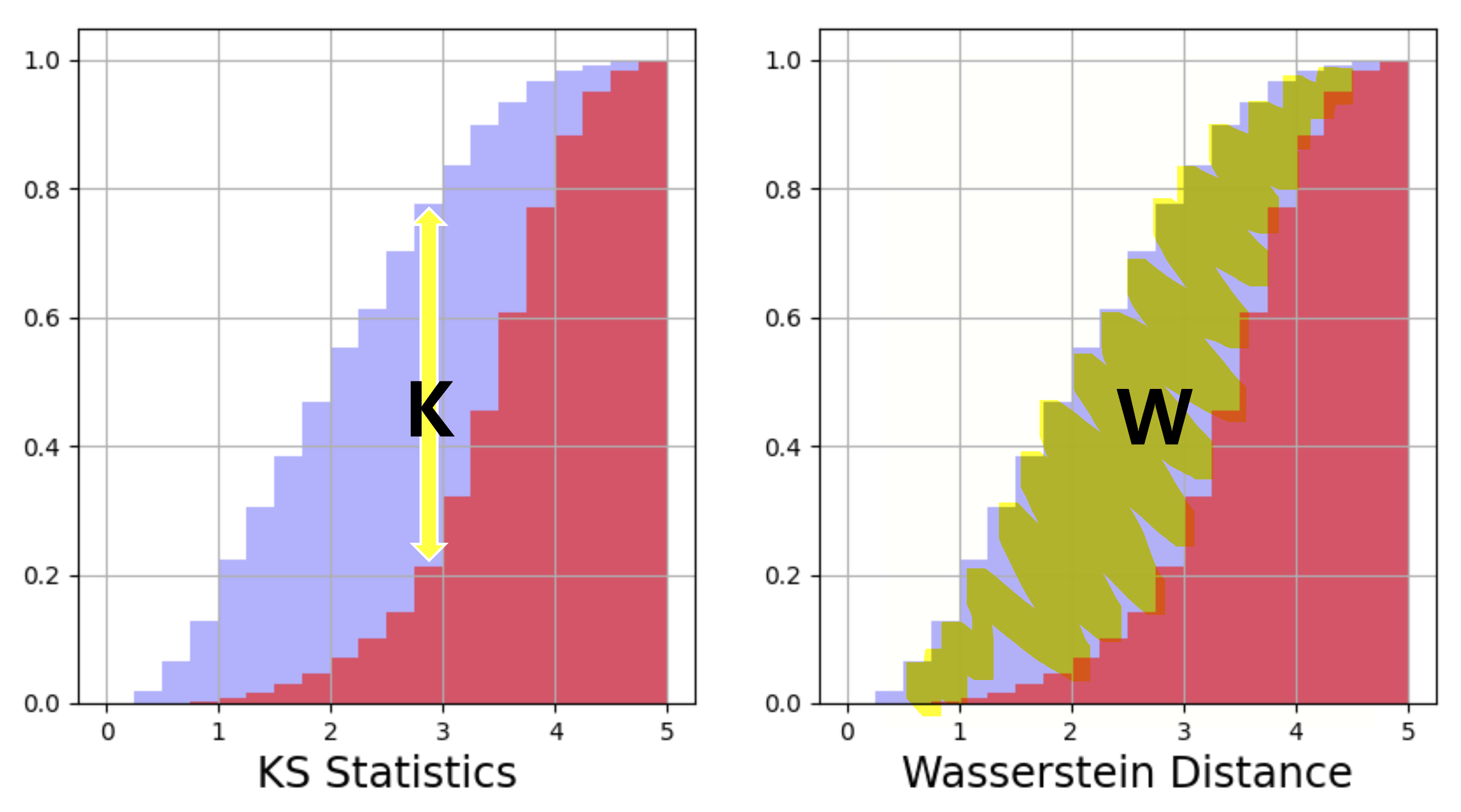}
    \caption{Visualization of the Kolmogorov–Smirnov statistic and Wasserstein distance. Red and blue histograms are separate cdfs to compare and the yellow arrow and the area represent each statistic.}
    \label{fig:metrics}
\end{figure}

We visualize two statistics, the KS statistic and Wasserstein distance for one-dimensional cdfs in \Cref{fig:metrics}.

\section{Basic Statistics of Data}
\subsection{Hallucination Data}
\label{apdx:stat_hallu}

On \Cref{tab:stat_detect}, we report the basic statistics of data utilized in \Cref{sec:distinguishability}.

\begin{table}[h]
    \centering
    \begin{tabular}{c|c|c|c}
        \toprule
        Data Set & Train & Valid (Dev) & Test \\
        \midrule
        \multirow{2}{*}{TC} & X & 383 & 3,845 \\
        ~ & ~ & \std{(305.11)} & \std{(301.49)} \\
        \midrule
        \multirow{2}{*}{WOW} & X & 430 & 3,607 \\
        ~ & ~ & \std{(55.09)} & \std{(55.19)} \\
        \midrule
        \multirow{2}{*}{CMU} & X & 416 & 3,607 \\
        ~ & ~ & \std{(225.68)} & \std{(226.14)} \\
        \midrule
        \multirow{2}{*}{FaithDial} & 33,887 & 6,297 & 6,441 \\
        ~ & \std{(110.18)} & \std{(112.04)} & \std{(110.41)} \\
        \midrule
        \multirow{2}{*}{XSUM} & X & 996 & 996 \\
        ~ & ~ & \std{(385.06)} & \std{(420.22)} \\
        \midrule
        \multirow{2}{*}{WikiBio} & X & 954 & 954 \\
        ~ & ~ & \std{(295.84)} & \std{(262.50)} \\
        \bottomrule
    \end{tabular}
    \caption{The number of data points and the average number of words of each data set.}
    \label{tab:stat_detect}
\end{table}

\subsection{Weighted Training Data}
\label{apdx:stat_train}

On \Cref{tab:stat_train} we report the basic statistics of data utilized on \Cref{sec:weighting}.

\begin{table}[h]
    \centering
    \begin{tabular}{c|c|c|c}
        \toprule
        Data Set & Train & Valid & Test \\
        \midrule
        \multirow{2}{*}{WOW} & 41,489 & 2,294 & 2,224 \\
        ~ & \std{(97.15)} & \std{(96.78)} & \std{(95.55)} \\
        \midrule
        \multirow{2}{*}{FaithDial} & 33,887 & 6,297 & 6,441 \\
        ~ & \std{(110.18)} & \std{(112.04)} & \std{(110.41)} \\
        \midrule
        \multirow{2}{*}{MediQA} & 578 & 29 & 45 \\
        ~ & \std{(334.82)} & \std{(447.41)} & \std{(509.77)} \\
        \bottomrule
    \end{tabular}
    \caption{The number of data points and the average number of words of each data set.}
    \label{tab:stat_train}
\end{table}

\section{Log Token Probability Distribution for Hallucinated and Entailed Data Sets}
\label{apdx:logtokenprobability}

\begin{figure*}[h]
    \centering
    \includegraphics[width=1.0\linewidth]{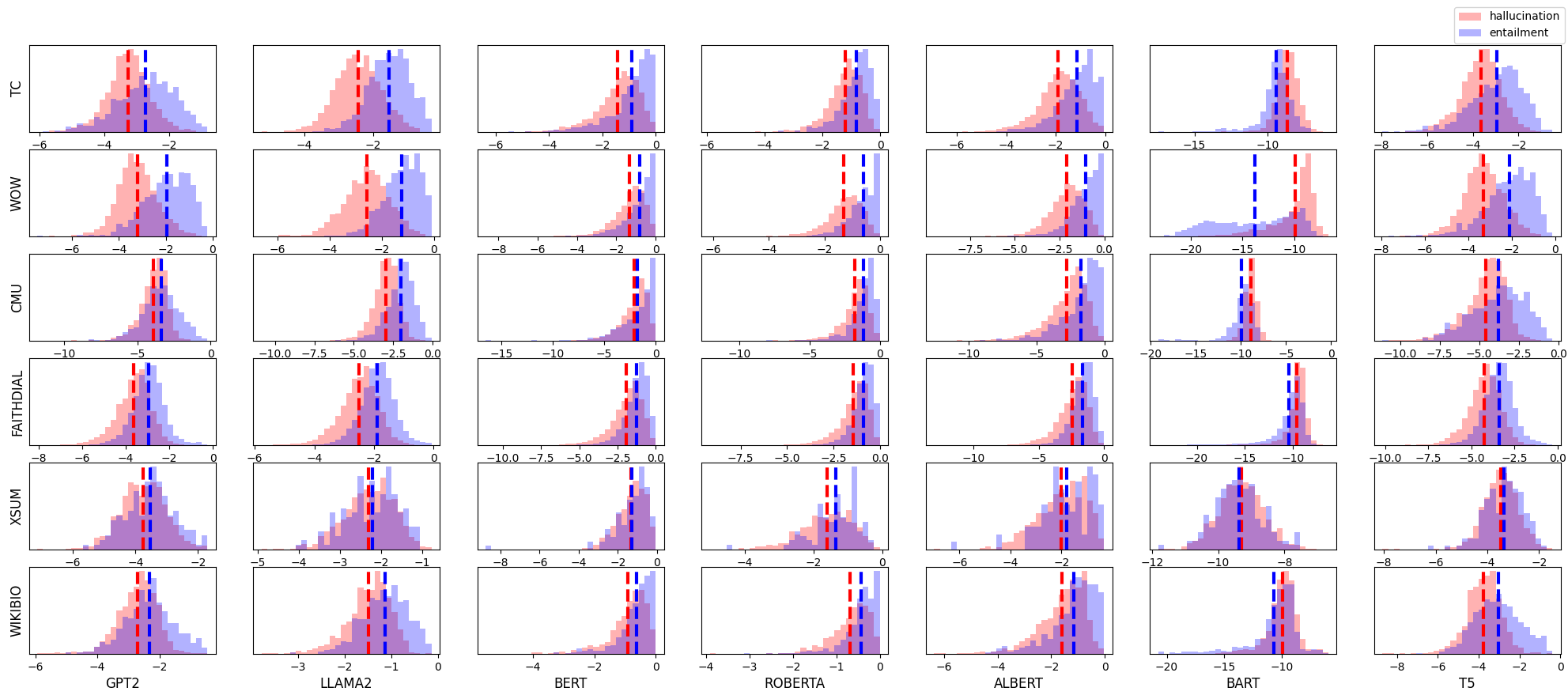}
    \caption{Empirical log token probability distribution and mean of $D_{Hallucinated}$ and $D_{Entailed}$ for each model and data set.}
    \label{fig:type_logits}
\end{figure*}

We present the log token probability distribution of $D_{Hallucinated}$ and $D_{Entailed}$ on \Cref{fig:type_logits}.
Probability on $D_{Hallucinated}$ is relatively lower than $D_{Entailed}$, except BART.
Roughly speaking, PLMs are internally more confident when they predict entailed texts.

\section{Fine-tuning Effect on CMU Data Set}
\label{apdx:cmu}

\begin{figure}[h]
    \centering
    \includegraphics[width=1.0\linewidth]{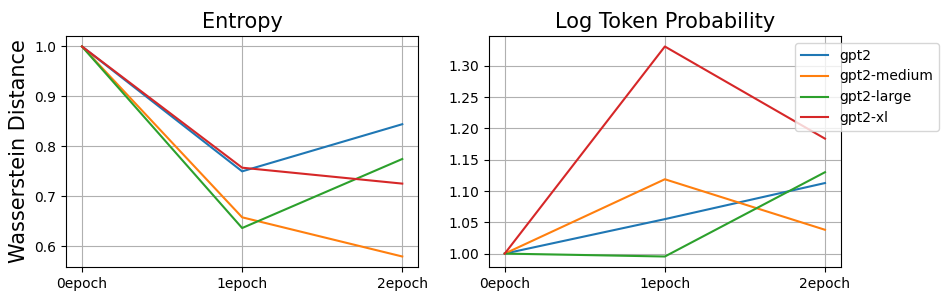}
    \caption{Fine-tuning effect for CMU data set. We divide all the Wasserstein distances with the statistics from the pre-trained model to visualize the relative change as training proceeds.}
    \label{fig:training_effect_cmu}
\end{figure}

We visualize the fine-tuning effect on the CMU data set on \Cref{fig:training_effect_cmu}.
We can check a similar trend with the WOW data set shown in \Cref{subsec:training_effect}.
Distinguishability with respect to entropy decreases while log token probability tends to increase.

\section{Experimental Detail for Weighted Training}
\label{apdx:experiment_detail}

\Cref{alg:weighted} depicts the algorithm in detail.
In lines 4 and 6, we compute each metric as proposed in \Cref{sec:distinguishability}.
In line 10, we apply the softmax function and multiply $N$ to obtain the same scale of total loss as the unweighted loss.

\begin{algorithm}[h]
   \caption{Weighted Training}
   \label{alg:weighted}
\begin{algorithmic}[1]
    \State {\bfseries Input:} Training data set $\mathcal{D}=\{(x_i,y_i)\}_{i=1}^N$, Target model $f$, Pre-trained reference model $g$, Target metric $M\in\{\text{Entropy, LogProb}\}$, Weight vector $W=\phi$
    \For{$i=1$ \textbf{to} $N$}
    \If{$M=$ Entropy}
        \State $w_i = - M(g(x_i, y_i))$
    \ElsIf{$M=$ LogProb}
        \State $w_i = M(g(x_i, y_i))$
    \EndIf
    \State $W \xleftarrow{} W \bigcup \{w_i\}$
    \EndFor
    \State $W \xleftarrow{}$ SoftMax$(W) \times N$
    \State \textbf{Train} $f$ with $w_i$Loss$(x_i,y_i)$
\end{algorithmic}
\end{algorithm}

Here are the detailed explanation on used baselines:

\begin{itemize}
    \item \textbf{Unweighted}: Usual unweighted training. It is equivalent to \Cref{alg:weighted} with $W=\mathbf{1_N}$.
    \item \textbf{CTRL}: \citet{rashkin-etal-2021-increasing} reported applying control tokens (<first-person>, <entailed>, <low-prec> etc.) improves faithfulness in knowledge-grounded dialogue. To obtain the control tokens, outer NER or NLI modules are required.\footnote{We utilize the CTRL implementation from https://github.com/McGill-NLP/FaithDial}
    \item Loss Truncation (\textbf{Truncation}): \citet{kang-hashimoto-2020-improved} suggested to truncate high-loss data points while training to achieve more faithful summarization. It is equivalent to \Cref{alg:weighted} if $w_i=1$ for low-loss data points and $w_i=0$ for the high-loss data points.\footnote{https://github.com/ddkang/loss\_dropper}
    \item \textbf{mFACT}: \citet{qiu-etal-2023-detecting} proposed another weighted-training method. They weigh the loss of each training example by its faithfulness score computed by a model trained on hallucination data sets. Though they proposed their method in a multi-lingual setting, it can be easily adapted to English.\footnote{https://huggingface.co/yfqiu-nlp/mFACT-en\_XX}
\end{itemize}

For the experiment, we use T5-small as $g$ and $f$ for our algorithm.
For baselines, we train T5-small (about 60 million parameters) with an AdamW \citep{loshchilov2018decoupled} optimizer with a learning rate of 1e-4 for all methods.
For the other hyper-parameters, we only utilize the default setting of packages and repositories.
We do not perform a hyper-parameter search.
We train each model 5 times and report the mean and standard deviation of each metric.
We use a machine with AMD Ryzen 9 5900X 12-Core Processor CPU with one NVIDIA RTX 3090 GPU.

Since our goal is to check the hallucination reduction performance of each training method, we only utilize training techniques from baselines, not decoding techniques (except attaching `<no-first-person> <entailed> <high-prec>' in CTRL decoding).
While decoding, we use greedy deterministic decoding to exclude external factors.

CTRL is specialized in the knowledge-grounded dialogue task while Truncation and mFACT are specialized in the summarization task.
So we use \textbf{WOW} \citep{dinan2019wizard} and \textbf{FaithDial} \citep{dziri-etal-2022-faithdial} as benchmarks for the knowledge-grounded dialogue task and \textbf{MediQA-AnS}\footnote{https://osf.io/fyg46/} \citep{savery2020question} for the summarization task.
Basic statistics of each data are reported on \Cref{apdx:stat_train}.

MediQA-AnS data set is a question-driven summarization data set consisting of (question, reference, answer (summary)) tuples.
The reference is a crawled web page and the answer is a human-written summary of the reference for the question.
MediQA-AnS targets consumer-level questions on healthcare information.
Since hallucinations in healthcare-related generations can severely harm human health, we select MediQA-AnS as a suited benchmark.
We use MediQA-AnS as a training set and use the NAACL-BioNLP 2021 - Task 2 data set as the test set, which covers the same task.\footnote{https://github.com/abachaa/MEDIQA2021/blob/main/ Task2/README.md}

For faithfulness evaluation, we utilize three metrics.
$\mathbf{Q^2}$ \citep{honovich-etal-2021-evaluating}\footnote{https://github.com/orhonovich/q-squared} first generates questions and answer candidates from the generated result.
Then $Q^2$ applies a QA model on the reference to solve the generated question.
After obtaining the answer from the reference, $Q^2$ compares it with the answer candidates lexically (\textbf{F1}) and semantically (\textbf{NLI}).
$Q^2$ has been utilized as a de facto method to measure hallucination.
Also, we use \textbf{SummaC} score \citep{laban-etal-2022-summac} and \textbf{FactKB}'s probability of entailment \citep{feng-etal-2023-factkb} as additional faithfulness metrics.

For general text quality evaluation, we utilize \textbf{ROUGE-L}\footnote{https://pypi.org/project/rouge/}, \textbf{BERTScore} \citep{Zhang*2020BERTScore:}\footnote{https://github.com/Tiiiger/bert\_score}, and \textbf{BARTScore} \citep{yuan2021bartscore}\footnote{https://github.com/stanfordnlp/string2string}.
We train each model 5 times and compute the mean and standard deviation of each score on the test set.

\end{document}